\documentclass{article}

\usepackage{arxiv}

\usepackage[utf8]{inputenc}
\usepackage[T1]{fontenc}
\usepackage{hyperref}
\usepackage{url}
\usepackage{booktabs}
\usepackage{amsfonts}
\usepackage{nicefrac}
\usepackage{microtype}
\usepackage{graphicx}
\usepackage{float}
\usepackage{siunitx}
\usepackage[numbers,compress]{natbib}
\usepackage{amsmath,amssymb}
\usepackage{bm,bbm}
\usepackage{multirow}
\usepackage{subcaption}
\usepackage{algorithm}
\usepackage{algpseudocode}

\graphicspath{{Figures/PDF/}{Figures/PNG/}}

\title{Deep Spatially-Regularized and Superpixel-Based Diffusion Learning for Unsupervised Hyperspectral Image Clustering}

\author{
  Vutichart Buranasiri \\
  Department of Mathematics\\
  Tufts University\\
  Medford, MA 02155, USA \\
  \And
  James M.\ Murphy \\
  Department of Mathematics\\
  Tufts University\\
  Medford, MA 02155, USA
}

\begin{document}
\maketitle

\begin{abstract}
An unsupervised framework for hyperspectral image (HSI) clustering is proposed that incorporates masked deep representation learning with diffusion-based clustering, extending the \textit{Spatially-Regularized Superpixel-based Diffusion Learning ($S^2DL$)} algorithm. Initially, a denoised latent representation of the original HSI is learned via an unsupervised masked autoencoder (UMAE) model with a Vision Transformer backbone. The UMAE takes spatial context and long-range spectral correlations into account and incorporates an efficient pretraining process via masking that utilizes only a small subset of training pixels. In the next stage, the entropy rate superpixel (ERS) algorithm is used to segment the image into superpixels, and a spatially regularized diffusion graph is constructed using Euclidean and diffusion distances within the compressed latent space instead of the HSI space. The proposed algorithm, \textit{Deep Spatially-Regularized Superpixel-based Diffusion Learning ($DS^2DL$)}, leverages more faithful diffusion distances and subsequent diffusion graph construction that better reflect the intrinsic geometry of the underlying data manifold, improving labeling accuracy and clustering quality. Experiments on Botswana and KSC datasets demonstrate the efficacy of $DS^2DL$.
\end{abstract}


\section{Introduction}

Machine learning methods have led to great developments in remote sensing \cite{wang2022self, li2024deep, ahmad2025comprehensive}, but many recent methods require a significant number of labeled training samples. Recent algorithms that employ a supervised deep learning architecture are capable of achieving high levels of labeling accuracy on benchmark hyperspectral images (HSI) \cite{upvehu_hyperspectral_scenes}. However, due to the large number of training labels that are required, the practicality of these algorithms is limited when labeled training data is not readily available. In response, unsupervised methods have been proposed that require no prior training labels, such as band-subset clustering \cite{zhao2011bandsubset}.

Recently, graph-based diffusion methods have been proposed, which cluster HSI by exploiting the intrinsic geometry of the dataset whilst maintaining efficient computation times \cite{murphy2018unsupervised, maggioni2019learning, murphy2019spectral, polk2021multiscale, polk2022unsupervised, polk2023unsupervised, Cui2024S2DL, tasissa2021deep}. In particular, the \emph{Superpixel-based Spatially regularized Diffusion Learning ($S^2DL$)} algorithm has shown superior labeling accuracy compared to other unsupervised methods on certain HSI datasets \cite{Cui2024S2DL}. Additionally, several supervised clustering algorithms have been proposed that utilize deep learning architectures such as convolutional neural networks \cite{hu2015deep, yu2017convolutional, cai2020graph, yang2017learning}, recurrent neural networks \cite{tulczyjew2020unsupervised}, and transformer-based models \cite{hong2021spectralformer, luo2024sdst}, which focus primarily on optimizing the HSI compression stage. While 3D-CNN models capture spatial context very well, they are unable to capture long-range spectral dependencies well like transformer-based models \cite{bengio1994learning}. The Masked Autoencoding Spatial-spectral Transformer (MAEST) algorithm utilizes an underlying supervised transformer-based architecture that captures spatial context to achieve superior labeling accuracy \cite{ibanez2022maest}.

Motivated by these recent developments, we propose the Deep Spatially-Regularized Superpixel-based Diffusion Learning ($DS^2DL$) algorithm that utilizes a compressed representation of the original HSI through an unsupervised variant of MAEST to extend the $S^2DL$ algorithm. In particular, the Unsupervised Masked Autoencoder (UMAE) first learns a denoised latent representation through an efficient pretraining process that requires a small subset of training pixels and accounts for spatial context and long-range spectral dependencies. Then, the entropy rate superpixel (ERS) algorithm is used to segment the image into superpixels, and the spatially regularized diffusion graph is constructed using diffusion and Euclidean distances within the latent space extracted from the UMAE. This ensures that the underlying diffusion graph encodes relationships between pixels that are unskewed by noise and spectral redundancy. Experiments on the Botswana and KSC HSIs demonstrate the effectiveness of $DS^2DL$ and show notable improvements in labeling accuracy, clustering quality and computational efficiency over $S^2DL$.

The remainder of this paper is organized as follows. In Section \ref{sec:Background}, we review the mathematical background of the MAEST and $S^2DL$ algorithms. Then, we describe the $DS^2DL$ algorithm in Section \ref{sec:Algorithm} and detail the experimental results in Section \ref{sec:Results}. Code is available at: \url{https://github.com/vburan01/DS2DL/tree/main}. 

\section{Background}
\label{sec:Background}

We first provide an overview of the prerequisite details regarding the HSI compression and HSI classification stages. The HSI compression stage incorporates the Reconstruction Encoder (RE) and Decoder (RD) from \cite{ibanez2022maest}. The HSI classification stage uses the HSI compression output to extend $S^2DL$.

\medskip
\noindent\textbf{Deep Learning for HSI Compression:}
Let $X \in \mathbb{R}^{H \times W \times B}$ denote a HSI of spatial dimension $H \times W$ with $B$ spectral bands. Let $p_i \in \mathbb{R}^B$ denote a pixel with $B$ bands and define the $p \times p$ patch $P_i \in \mathbb{R}^{p \times p \times B}$ with central pixel $p_i$. $P_i$ contains the spectral signatures of all $p^2$ pixels in the spatial neighborhood. For each band $b$, define $P_b^{(i)}\in \mathbb{R}^{p^2 \times 1}$ to be the flattened horizontal slice of $P_i$ containing the $b^{\text{th}}$ spectral band of each pixel in $P_i$. Define the spatial-spectral group $G_b^{(i)}\in \mathbb{R}^{\ell p^2 \times 1}$ of pixel $p_i$ centered at band $b$, where

\[
G_b^{(i)} = [P_{b - \lfloor\ell/2 \rfloor }^{(i)} ,  P_{b - \lfloor\ell/2 \rfloor +1 }^{(i)}, \dots , P_{b + \lfloor\ell/2 \rfloor }^{(i)}]^T .
\]

The parameter $\ell$ is the number of bands in each group $G_b^{(i)}$. If we have $P_r^{(i)}$ such that $r \notin \{1,\dots,B\}$, we wrap around the spectrum and define $\tilde{b} := ((r - 1)\bmod B) + 1$ and set $P_r^{(i)} := P_{\tilde{b}}^{(i)}$. Combining each $G_b^{(i)}$ for every $b \in B$ yields the matrix

\[
V_G^{(i)} = [ G_1^{(i)} \; G_2^{(i)} \cdots G_B^{(i)}] \in \mathbb{R}^{\ell p^2 \times B}.
\]

We introduce a learnable linear projection matrix $W \in \mathbb{R}^{D \times \ell p^2}$ that acts on each group $G_b^{(i)}$ in $V_G^{(i)}$, to produce a compressed $D$-dimensional representation of each group given by $E_b^{(i)} = WG_b^{(i)} \in \mathbb{R}^{D \times 1}$, where $D$ is a parameter denoting the latent dimension. We call each $E_b^{(i)}$ a groupwise token and obtain the matrix of groupwise tokens $M_E^{(i)} = WV_G^{(i)} \in \mathbb{R}^{D \times B}$, with columns $E_b^{(i)}$. We then define the Learnable Positional Encoding (LPE) matrix $s_G \in \mathbb{R}^{D \times B}$ where the $b^{\text{th}}$ column of $s_G$ is a band-specific vector that encodes learnable weights that are added to $E_b^{(i)}$. The model thus accounts for spectral position during the learning process. We then obtain the positional token matrix 

\[M_{pos}^{(i)} = WV_G^{(i)} + s_G \in \mathbb{R}^{D \times B}\]

For pretraining, we introduce a masking operator $g(\cdot)$, which randomly removes $R_m\%$ of positional tokens corresponding to groups in $V_G^{(i)}$, where $R_m \in [0,1]$ denotes the masking ratio. This gives the matrix of unmasked positional tokens
\[
M^{(i)} = g(WV_G^{(i)} + s_G) \in \mathbb{R}^{D \times K},
\]
where $K = (1-R_m)B$ is the number of unmasked groups. The remaining masked groups are reconstructed by the RD later in the pipeline. The matrix $M^{(i)}$ consisting of the $K$ unmasked tokens is inputted into a Vision Transformer (ViT) encoder. The ViT encoder consists of a stack of transformer blocks with multi-head self-attention and Multilayer-Perceptron (MLP) layers. Further details on the ViT architecture are found in \cite{dosovitskiy2020image}. The ViT encoder outputs a sequence of contextualized latent tokens $L^{(i)} \in \mathbb{R}^{D \times K}$ per pixel, which capture the aggregated spatial-spectral information of the patch $P_i$ across its unmasked spectral groups. Finally, the per-pixel latent $L^{(i)}_{\text{mean}} \in \mathbb{R}^{D}$ is obtained by averaging the $K$ output tokens via mean-pooling.

The RD takes in the RE output $L^{(i)}$ for each $i$. We apply a linear projection layer $U \in \mathbb{R}^{D \times D}$ that maps each $L^{(i)}$ to the decoder projection space and define an analogous LPE matrix $s'_G \in \mathbb{R}^{D \times B}$ for the decoder over all bands $b$. Adding the columns of $s'_G$ for each visible token in the matrix $L^{(i)}$ yields the unmasked positional tokens $Z_{vis}^{(i)} \in \mathbb{R}^{D \times K}$. We then define a learnable masked token $m \in \mathbb{R}^{D}$. For each masked band $b$, we add the positional encoding of $b$ to $m$, and concatenate these vectors to form the matrix $m_{pos} \in \mathbb{R}^{D \times (B-K)}$.

Concatenating $m_{pos}$ and $Z_{vis}^{(i)}$ yields the decoder input sequence $Z^{(i)} \in \mathbb{R}^{D \times B}$, which is passed through a stack of ViT decoder blocks to produce the matrix of outputted tokens $\tilde{Z}^{(i)} \in \mathbb{R}^{D \times B}$. Then, a final linear projection layer $H \in \mathbb{R}^{\ell p^2 \times D}$ is applied to each of the masked tokens of $\tilde{Z}^{(i)}$ to produce the $(B-K)$ reconstructed masked spatial-spectral groups $\tilde{G}_b^{(i)} \in \mathbb{R}^{\ell p^2 \times 1}$. The training objective is to minimize the mean squared error (MSE) loss between $\tilde{G}_b^{(i)}$ and $G_b^{(i)}$ for each masked band $b$. The learnable parameters in $W, s_G, U, s'_G, m, H$ and parameters in the ViT encoder and decoder are optimized jointly via backpropagation.

\vspace{10pt}
\noindent\textbf{Diffusion Learning for HSI Clustering:}
The $S^2DL$ algorithm starts by applying Principal Component Analysis (PCA) on the flattened HSI, $X_{flat} \in \mathbb{R}^{HW \times B}$, to reduce $X$ to its first 3 Principal Components (PCs), yielding $X_{PC} \in \mathbb{R}^{H \times W \times 3}$. The Entropy Rate Superpixel (ERS) algorithm \cite{liu2011entropy} is run on $X_{PC}$ to segment the image into $N_s$ superpixels, where $N_s$ is the number of superpixels. A diffusion process is run on the space of superpixels to segment the HSI as follows. Let $x \in \mathbb{R}^{B}$ denote a pixel. Let $k_n(x)$ denote the $k$ nearest neighbors of $x$ under the $\ell^2$ metric. The local density $\zeta(x)$ of $x$ is given by
\[
\zeta(x) = \sum_{y \in k_n(x)} \exp(-\| x-y\|_2^2/\sigma_0^2),
\]

where $\sigma_0>0$ is a scaling parameter controlling interaction radius between pixels. We define the set

\[
X_s = \bigcup_{i=1}^{N_s} \{x \in S_i: x \text{ is one of the } k \text{ maximizers of } \zeta(x)\},
\]

where $S_i$ denotes a superpixel. The $S^2DL$ pipeline constructs a spatially regularized kNN graph over the $kN_s$ pixels in $X_s$, represented by the weighted adjacency matrix $\mathbf{W} \in \mathbb{R}^{kN_s \times kN_s}$. We set

\[
\mathbf{W}_{ij} = 
\begin{cases} 
\exp(-\|x_i-x_j\|_2^2/\sigma_0^2 ) &   \|(h_{x_i}, w_{x_i})- (h_{x_j},w_{x_j})\|_2 \le R, \\[4pt] 
0, & \text{Otherwise,}
\end{cases}
\]

where $(h_{x_i}, w_{x_i})$ are the spatial coordinates of $x_i \in X$ and $R$ is the spatial radius parameter, and let $\mathbf{W}_{ij} = 0$ otherwise.

Diffusion distances \cite{coifman2005geometric, coifman2006diffusion} are calculated through a Markov diffusion process on $\mathbf{W}$. We define the diagonal degree matrix $\mathbf{D}$ by $\mathbf{D}_{ii} = \sum_{j=1}^N \mathbf{W}_{ij}$, and form the transition matrix $\mathbf{P}$ by $\mathbf{P} = \mathbf{D}^{-1}\mathbf{W}$. Then, given that $\mathbf{P}$ is irreducible and aperiodic, we have that $\mathbf{P}$ has a unique stationary distribution $\pi \in \mathbb{R}^{kN_s}$ such that $\pi \mathbf{P} = \pi$. We denote the eigenvector-eigenvalue pairs of $\mathbf{P}$ by $\{(\lambda_m, \psi_m)\}_{m=1}^{kN_s}$, and define the diffusion distance of $x_i$ and $x_j$ in $X$ at a time $t \ge 0$ by

\[
D_t(x_i,x_j) = \sqrt{\sum_{m=1}^{kN_s} [ (\mathbf{P}^t)_{im} - (\mathbf{P}^t)_{jm}]^2/\pi_m}
= \sqrt{\sum_{m=1}^{kN_s} |\lambda_m|^{2t} [(\psi_m)_i - (\psi_m)_j]^2 }.
\]

Then for a pixel $x \in X$, define

\[
d_t(x) =
\begin{cases}
\max_{y \in X_s } D_t(x,y), & x = \arg \max_{y \in X_s} \zeta(y), \\[4pt]
\min_{y \in X_s} \{D_t(x,y) : \zeta(y) \ge \zeta(x) \}, & \text{otherwise,}
\end{cases}
\]

and denote the mode score of a pixel $x\in X$ by $\Delta_t(x) = \zeta(x)d_t(x)$. In particular, $S^2DL$ picks out the $K$ maximizers of $\Delta_t(x)$, which are pixels that are highest in local density and farthest away in diffusion distance from other higher density pixels, which we call cluster modes. Given a cluster mode $x$, we denote its Local Backbone (LBB) as
\[ 
LBB(x) = \{x\} \cup \{k_n \text{ nearest neighbors of } x\} 
\]
$S^2DL$ hierarchically assigns labels: first, all $K$ modal pixels are assigned a unique label; then, for each modal pixel $x$, all pixels in $LBB(x)$ are assigned the same label as $x$. Then, in decreasing order of $\zeta(x)$, the remaining unlabeled pixels in $X_s$ containing $x$ are assigned the label of the $D_t$ nearest neighbor that is already labeled and has greater density. Finally, majority voting is used to designate labels to the remaining pixels in each superpixel $S_i$, assigning the majority label of the $k$ representatives in $S_i$ to these pixels.

\section{The Proposed Algorithm}
\label{sec:Algorithm}

The proposed algorithm, $DS^2DL$, uses a learned latent representation of the original HSI data to improve the diffusion learning algorithm by removing spectral noise, leading to higher downstream classification accuracy. First, the Unsupervised Masked Autoencoder (UMAE) learns a compressed latent featuring only spectrally significant HSI signatures per pixel. Second, a modified version of $S^2DL$ is run, where the weighted adjacency matrix $\mathbf{W}$ and corresponding transition matrix $\mathbf{P}$ are constructed using spectral data from the learned latent space $L$, rather than $X$.

The UMAE is outlined as follows in Algorithm~\ref{alg:uMVAE}, and produces the compressed latent $L \in \mathbb{R}^{H \times W \times D}$. In our implementation, we fix $D = 48$. We first flatten the HSI $X$ and project $X$ onto its first 20 PCs to reduce spectral redundancy. Farthest Point Sampling (FPS) \cite{eldar1997farthest} is used to pick $n_t$ training pixels, which iteratively chooses the $(k+1)^{\text{th}}$ pixel with the greatest minimum difference in $\ell^2$ spectral norm to the first $k$ training pixels. FPS ensures a spectrally diverse set of training pixels while remaining computationally efficient and completely unsupervised.

During pretraining, we extract the $p \times p$ spatial patches of each $i^{\text{th}}$ training pixel which encode local spatial context in addition to spectral information. The construction of the spatial-spectral groups $G_b^{(i)}$ leverages the notion that discriminative spectral information is concentrated in adjacent spectral bands and spatially adjacent pixels. Additionally, incorporating the learnable positional encodings $s_G$ allows the AE to learn the ordered structural context of HSI spectra.

The benefits of incorporating masking are twofold: increasing $R_m\%$ makes the reconstruction task more difficult and forces the model towards spectrally rich and meaningful latent representations. Additionally, focusing the reconstruction task to $R_m\%$ of spatial-spectral groups per training patch ensures optimal and efficient pretraining. Furthermore, the underlying ViT architecture encourages the model to learn correlations between spectral bands that are further apart in wavelength. Hence, the proposed algorithm learns representations that leverage local spatial data and long-range spectral information.

In the second stage, $DS^2DL$ utilizes the learned latent $L$ to compute the adjacency matrix $\mathbf{W}$: given the HSI compressed spectra of $x_i$ under $L$, denoted $x'_i \in \mathbb{R}^{D}$, we set
\[
\mathbf{W}_{ij} = 
\begin{cases} 
\exp(-\|x'_i-x'_j\|_2^2/\sigma_0^2 ) &   \|(h_{x_i}, w_{x_i})- (h_{x_j},w_{x_j})\|_2 \le R, \\[4pt] 
0, & \text{Otherwise,}
\end{cases}
\]
The corresponding transition matrix $\mathbf{P}$ and diffusion distances $D_t$ are calculated under $L$ as opposed to $X$, which further significantly reduces runtime.

\begin{algorithm}[H]
  \caption{Unsupervised Masked Autoencoder (UMAE)}
  \label{alg:uMVAE}
  \begin{algorithmic}[1]
    \Statex
    \State \textbf{Input}: $X \in \mathbb{R}^{H \times W \times B}$ (HSI), $n_t$ (\# of training pixels): $R_m$ (Mask Ratio), $p$  (Patch Size), $\ell$ (Band Neighbor Length), $D$ (Latent Dimension), $N_e$ (\# Training Epochs) 
    \State Normalize spectral bands of $X$ to $[0,1]$  and create $X_{flat}\in \mathbb{R}^{H W \times B}$ and the $p \times p$ padded HSI $\hat{X}$.
    \State Reduce $X_{flat}$ to 20 bands using PCA and run FPS to select a set of $n_t$ training pixels, denoted $S$. 
      \State \textbf{For each $i \in S$:}
      \State Extract $P_i \in \mathbb{R}^{p \times p\times B}$ from $\hat{X}$ and find $G_b^{(i)}$ for each band.
      \State Obtain $V_G^{(i)}$ and apply $W\in \mathbb{R}^{D \times \ell p^2 }$ and $s_G( \cdot )$ to obtain the matrix of tokens $M_{pos}^{(i)}$.
      \State Apply $g(\cdot)$, masking $R_m \% $ of groups in $M_{pos}^{(i)}$ to get $M^{(i)}$.
      \State \textbf{For $t = 1 , \dots, N_e$ Epochs:} 
      \State Pretrain the ViT AE to reconstruct spatial-spectral groups with $M^{(i)}$ as input for all $i \in S$ with MSE.
      \State \textbf{For each pixel $p_i \in X$:}
      \State Repeat steps 5 and 6 to obtain $M^{(i)}$, then run ViT encoder unmasked with $R_m = 0$ to obtain per-pixel latent representation $L^{(i)} \in \mathbb{R}^{D \times B} $
      \State Use mean-pooling to obtain pixel features $L_{mean}^{(i)} \in \mathbb{R}^{D}$
      \State Concatenate each $L_{mean}^{(i)}$ and reshape into latent representation $L \in \mathbb{R}^{H \times W \times D}$
    \State \textbf{Output:} learned latent representation $L \in \mathbb{R}^{H \times W \times D}$
  \end{algorithmic}
\end{algorithm}

\begin{algorithm}[H]
  \caption{$DS^2DL$ Algorithm}
  \label{alg:s2dl_latent}
  \begin{algorithmic}[1]
    \Statex
    \State \textbf{Input}: $X \in \mathbb{R}^{H \times W \times B}$ (HSI), $L \in \mathbb{R}^{H \times W \times D}$ (latent feature map), 
           $N_s$ (\# of superpixels), $k$ (\# of representative pixels per superpixel), 
           $\sigma_0$ (kernel scaling factor), $k_n$ (\# of nearest neighbors), 
           $R$ (spatial radius), $K$ (\# of clusters).

    \State Flatten $X$ into a matrix $X_{flat} \in \mathbb{R}^{HW \times B}$ and project $X_{flat}$ onto its first 3 PCs and reshape to obtain $X_{\mathrm{PC}}$.
    \State Run ERS on $X_{\mathrm{PC}}$ to segment the image into $N_s$ superpixels.
    \State For each pixel $p_i$, find its $k$ nearest
           neighbors $k_n(x)$ in the latent space $L$ and compute 
           $\zeta(x)$.
    \State For each superpixel, select the $k$ pixels with largest $\zeta(x)$ and collect them 
           into the representative set $X_s$.
    \State Construct a spatially-regularized $kNN$ graph over $X_s$ with $k \cdot N_s$ pixels using radius $R$.
    \State Compute diffusion distances $d_t(x)$ on the graph for all $x \in X_s$ and form 
           $\Delta_t(x) = \zeta(x)\,d_t(x)$.
    \State Find the $K$ maximizers of $\Delta_t(x)$ and assign them 
           unique labels $1,\dots,K$.
    \State For each modal pixel, assign the its corresponding LBB with $k_n$ neighbors the same label as itself. 
    \State In order of decreasing density $\zeta(x)$, assign each remaining unlabeled pixel 
           in $X_s$ the label of its $D_t$ nearest neighbor with higher density.
    \State For each superpixel, assign all pixels in that superpixel the modal label among the 
           $k$ representative pixels selected from it, obtaining the final clustering map $C$.
    \State \textbf{Output}: clustering map $C \in \{1,\dots,K\}^{H \times W}$.
  \end{algorithmic}
\end{algorithm}

\section{Experimental Results}
\label{sec:Results}

We demonstrate the improved experimental accuracy and efficiency of $DS^2DL$ over $S^2DL$ on the Botswana and Kennedy Space Center (KSC) \cite{upvehu_hyperspectral_scenes} HSIs. The Botswana scene was captured by the NASA EO-1 Satellite over the Ovakango Delta, Botswana, in 2004, with a 30-meter spatial resolution per pixel and 145 spectral bands. The GT has 14 classes and has a scene size of $1476 \times 256$ pixels. The KSC scene was captured by the AVIRIS sensor over wetland areas of NASA's Kennedy Space Center, Florida in March 1996, with an 18-meter spatial resolution per pixel and 176 spectral bands. The GT has 13 classes and has a scene size of $512 \times 614$ pixels.

We evaluate the performance of $S^2DL$ and $DS^2DL$ using Overall Accuracy (OA, total fraction of correctly labeled pixels), Average Accuracy (AA, mean per-class accuracies), Cohen's Kappa ($\mathcal{K}$), purity with different numbers of clusters used ($1\times$, $2\times$, $3\times$ number of ground truth classes), and Normalized Mutual Information (NMI). The main input parameters in $S^2DL$ and $DS^2DL$, as displayed in Algorithm~\ref{alg:s2dl_latent}, were optimized via a small grid search. Input parameters in Algorithm~\ref{alg:uMVAE} for $DS^2DL$ were optimized via a sequential 1D parameter search. OA, AA, and $\mathcal{K}$ were optimized jointly, and the remaining metrics were optimized independently. The runtime (RT) of the run maximizing OA, AA, and $\mathcal{K}$ is tracked.

\begin{table}[H]
\centering
\small
\caption{Comparison of S\textsuperscript{2}DL and DS\textsuperscript{2}DL across KSC and Botswana datasets.}
\begin{tabular}{llcc}
    \toprule
    \textbf{Dataset} & \textbf{Metric} & \textbf{S\textsuperscript{2}DL} & \textbf{DS\textsuperscript{2}DL} \\
    \midrule
    \multirow{7}{*}{KSC}
      & OA        & 0.56688 & \textbf{0.6008} \\
      & AA        & 0.52220 & \textbf{0.6247} \\
      & $\kappa$  & 0.54115 & \textbf{0.5618} \\
      & $1\times$ Purity & 0.6279  & \textbf{0.6824} \\
      & $2\times$ Purity & 0.7192  & \textbf{0.7972} \\
      & $3\times$ Purity & 0.8035  & \textbf{0.8428} \\
      & NMI       & 0.6766  & \textbf{0.7182} \\
      & RT (s)    & 2805.71 & \textbf{934.03} \\
    \midrule
    \multirow{7}{*}{Botswana}
      & OA        & 0.60037 & \textbf{0.64101} \\
      & AA        & 0.61575 & \textbf{0.66476} \\
      & $\kappa$  & 0.56781 & \textbf{0.61207} \\
      & $1\times$ Purity & 0.6176  & \textbf{0.6281} \\
      & $2\times$ Purity & 0.7087  & \textbf{0.7485} \\
      & $3\times$ Purity & 0.7657  & \textbf{0.8168} \\
      & NMI       & 0.7083  & \textbf{0.7244} \\
      & RT (s)    & 2782.62 & \textbf{947.83} \\
    \bottomrule
\end{tabular}
\label{tab:results}
\end{table}

\begin{figure}[H]
\centering
\begin{subfigure}[t]{0.11\textwidth}
  \includegraphics[width=\linewidth]{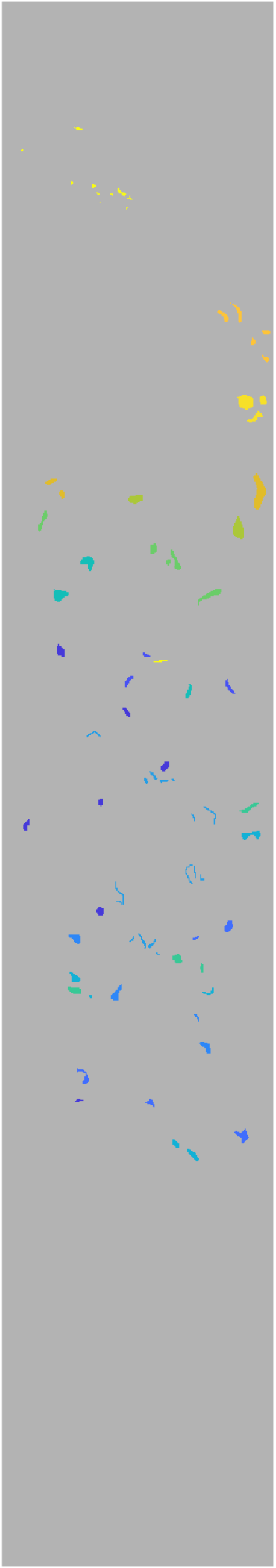}
  \caption{GT}
\end{subfigure}
\begin{subfigure}[t]{0.11\textwidth}
  \includegraphics[width=\linewidth]{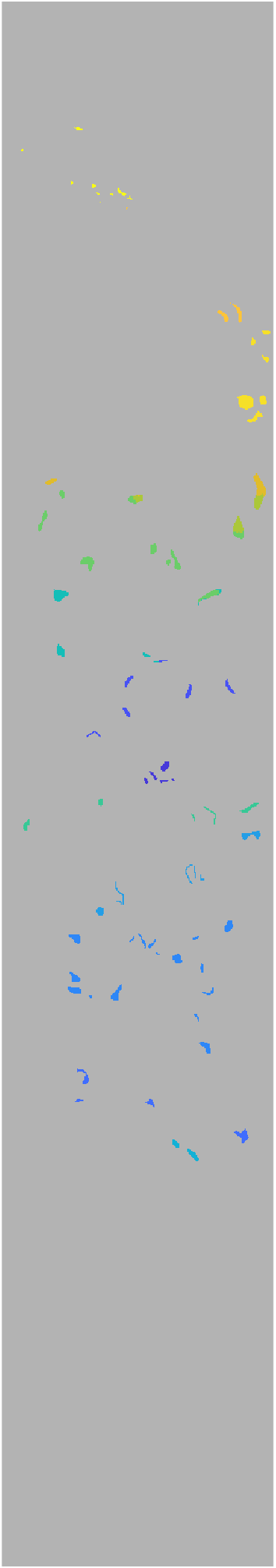}
  \caption{$S^2DL$}
\end{subfigure}
\begin{subfigure}[t]{0.11\textwidth}
  \includegraphics[width=\linewidth]{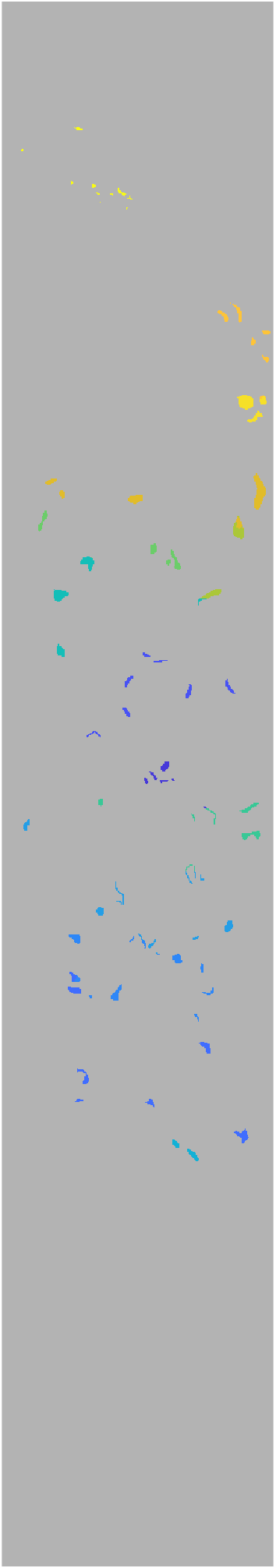}
  \caption{$DS^2DL$}
\end{subfigure}

\vspace{0.5em}

\begin{subfigure}[t]{0.15\textwidth}
  \includegraphics[width=\linewidth]{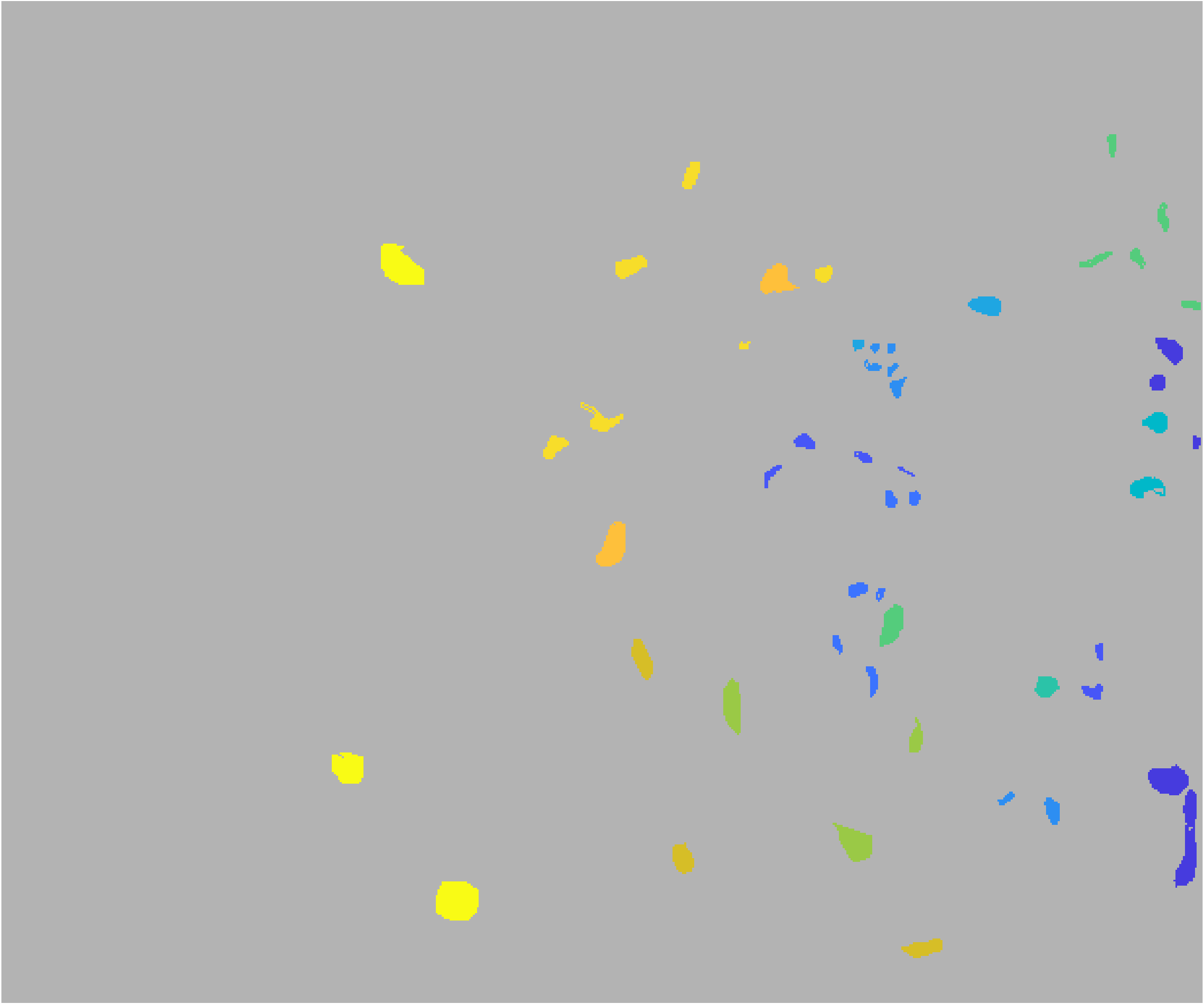}
  \caption{GT}
\end{subfigure}
\begin{subfigure}[t]{0.15\textwidth}
  \includegraphics[width=\linewidth]{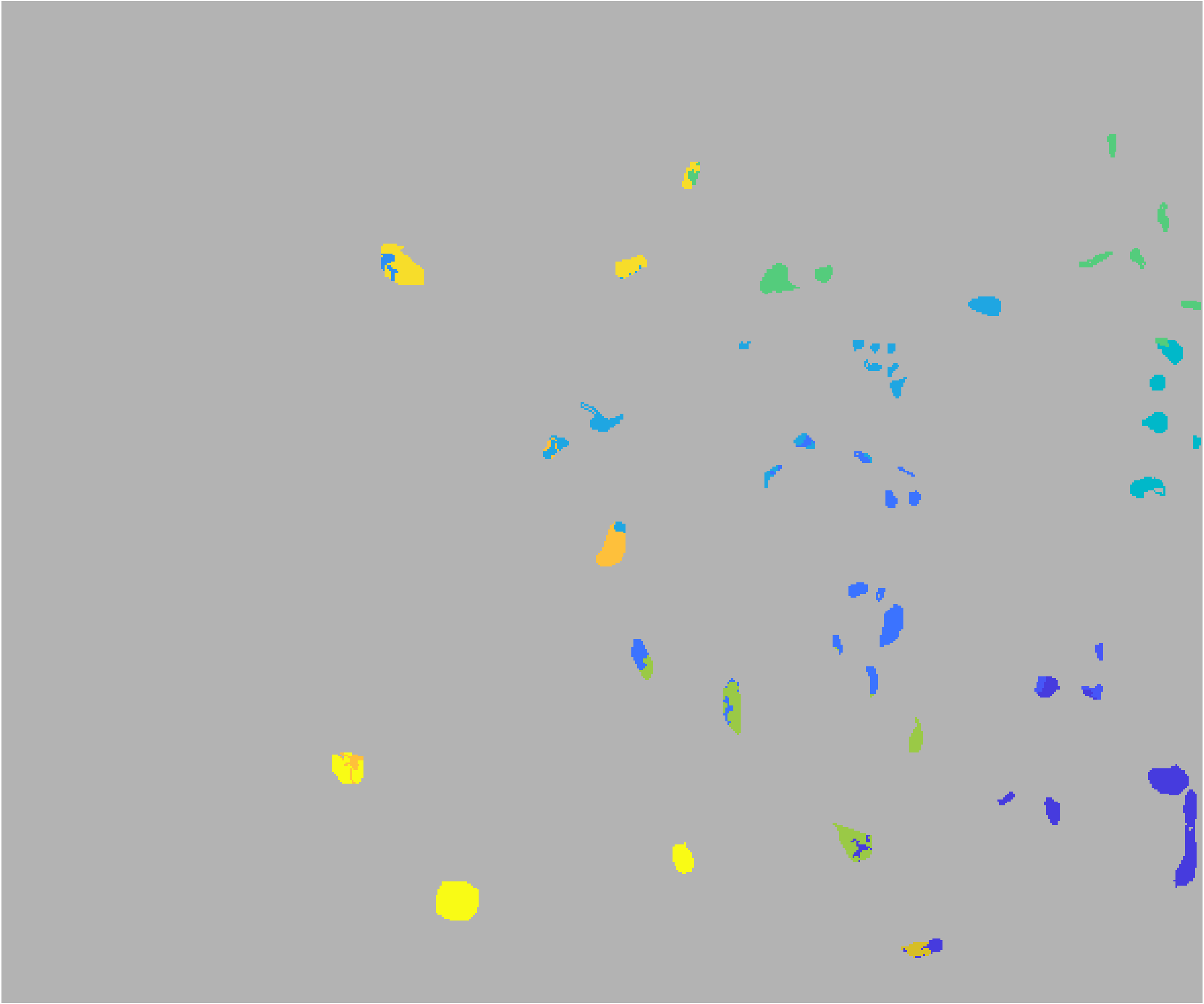}
  \caption{$S^2DL$}
\end{subfigure}
\begin{subfigure}[t]{0.15\textwidth}
  \includegraphics[width=\linewidth]{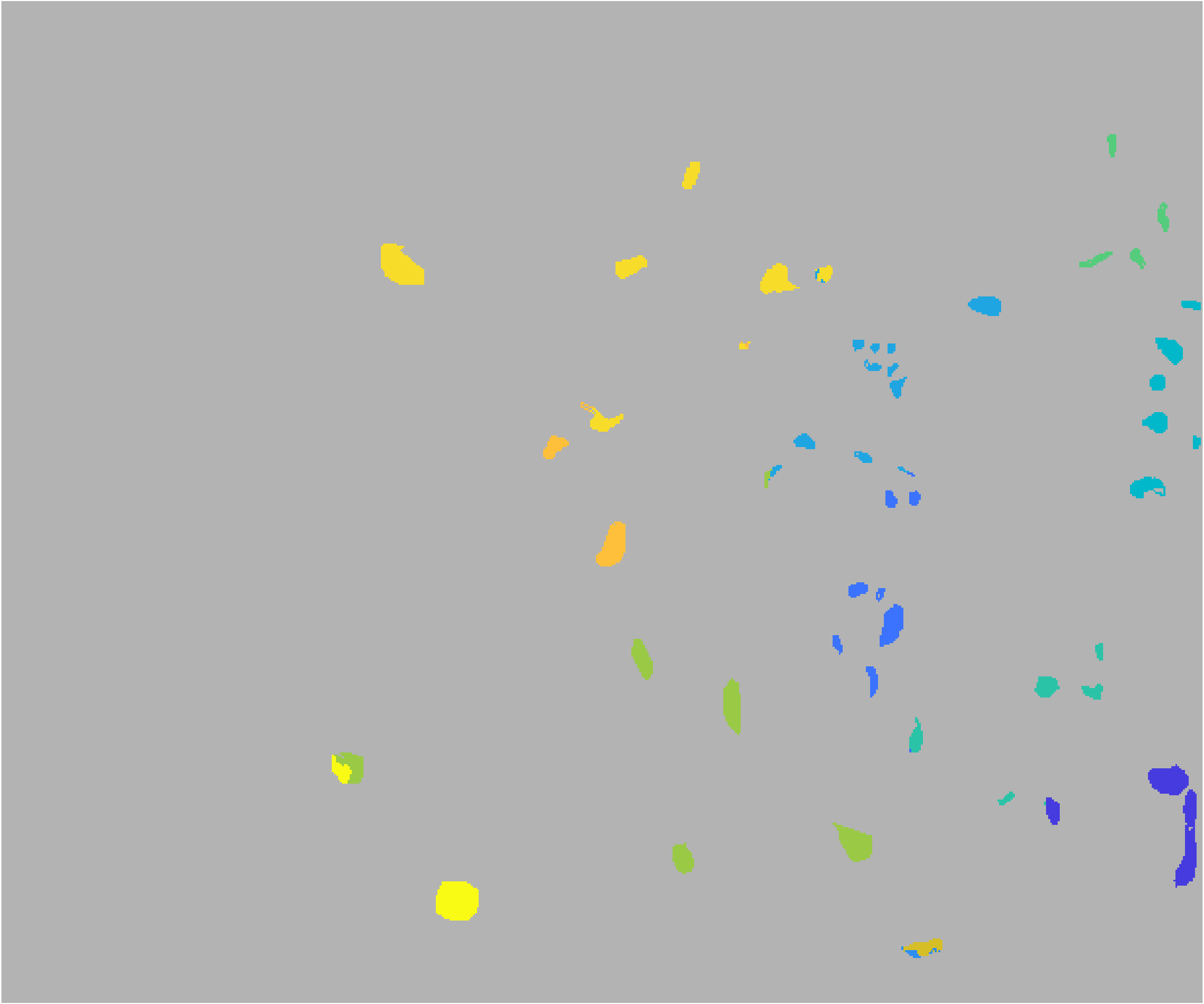}
  \caption{$DS^2DL$}
\end{subfigure}

\caption{Top: Botswana; Bottom: KSC.}
\label{fig:results}
\end{figure}

Table~\ref{tab:results} highlights the efficacy of the $DS^2DL$ algorithm over $S^2DL$ throughout all metrics discussed above. In the KSC dataset, we observe a cumulative increase of $0.37$ points over all metrics, indicating noticeable improvements in both accuracy and clustering quality. Most notably, the largest jump is in AA, where $DS^2DL$ boasts a $10\%$ improvement over $S^2DL$. $DS^2DL$ thus improves labeling accuracy in more difficult, underrepresented classes, where $S^2DL$ is limited. Substantial improvements in purity and NMI scores further reflect that $DS^2DL$ forms more internally coherent clusters with greater alignment to ground truth clusters.

The performance of $DS^2DL$ in the Botswana dataset indicates similar but more moderate improvements over $S^2DL$. $DS^2DL$ improves labeling accuracy and clustering quality similarly---all accuracy metrics, as well as $2\times$ and $3\times$ purity, improve by $4$--$5\%$, whilst NMI and $1\times$ purity improve by $1$--$2\%$. In both datasets, $DS^2DL$ boasts a significant reduction in RT, since calculating the $k$ nearest neighbors per pixel in the compressed latent space is substantially less computationally expensive than in the HSI space.
\section{Conclusions and Future Directions}

The proposed algorithm, $DS^2DL$, provides noticeable improvements over $S^2DL$ in benchmark accuracy and clustering quality metrics whilst significantly reducing runtime. In future work, we may seek to introduce a fine-tuning stage of the masked autoencoder model using unsupervised contrastive learning methods to further increase denoising and compression quality. Additionally, understanding the relationships between hyperparameters in the Diffusion Learning and HSI compression pipelines is a further topic of interest that may be explored more thoroughly. Extensions to semisupervised and active learning are also of interest \cite{murphy2018iterative, murphy2020spatially, polk2022active}.

\bibliographystyle{unsrt}
\bibliography{refs}

\end{document}